\documentclass[a4paper,10pt,twoside]{article}

\usepackage{clin}        % CLIN Journal style
\usepackage{harvard}     % Harvard bibliography style

\usepackage[T1]{fontenc}
\usepackage[utf8]{inputenc}
\usepackage{graphicx}
\usepackage{booktabs}
\usepackage{tabularx}
\usepackage{array}
\usepackage{amsmath}
\usepackage{multirow}
\usepackage[hidelinks]{hyperref}
\usepackage{url}

\newcolumntype{Y}{>{\raggedright\arraybackslash}X}

\newcommand{\figref}[1]{Fig.~\ref{#1}}
\newcommand{\tabref}[1]{Table~\ref{#1}}

\begin{document}

\title{Exploring Cultural Variations in Moral Judgments with Large Language Models}

\author{Hadi Mohammadi$^*$ \email{h.mohammadi@uu.nl}\\
{\normalsize \bf Ayoub Bagheri}$^*$ \email{a.bagheri@uu.nl}
\AND \addr{$^*$Department of Methodology and Statistics, Utrecht University, Utrecht, The Netherlands}}

\maketitle\thispagestyle{empty}

\begin{abstract}
Large Language Models (LLMs) have shown strong performance across many tasks, but their ability to capture culturally diverse moral values remains unclear. In this paper, we examine whether LLMs mirror variations in moral attitudes reported by the World Values Survey (WVS) and the Pew Research Center's Global Attitudes Survey (PEW). We compare smaller monolingual and multilingual models (GPT-2, OPT, BLOOMZ, and Qwen) with recent instruction-tuned models (GPT-4o, GPT-4o-mini, Gemma-2-9b-it, and Llama-3.3-70B-Instruct). Using log-probability-based \emph{moral justifiability} scores, we correlate each model's outputs with survey data covering a broad set of ethical topics. Our results show that many earlier or smaller models often produce near-zero or negative correlations with human judgments. In contrast, advanced instruction-tuned models achieve substantially higher positive correlations, suggesting they better reflect real-world moral attitudes. We provide a detailed regional analysis revealing that models align better with Western, Educated, Industrialized, Rich, and Democratic (W.E.I.R.D.) nations than with other regions. While scaling model size and using instruction tuning improves alignment with cross-cultural moral norms, challenges remain for certain topics and regions. We discuss these findings in relation to bias analysis, training data diversity, information retrieval implications, and strategies for improving the cultural sensitivity of LLMs.
\end{abstract}

\begin{keywords}
Large Language Models, Moral judgments, Cultural variation, Cross-cultural analysis, Bias detection
\end{keywords}

\section{Introduction}

Over the past few years, Large Language Models (LLMs) have gained prominence in both academic and public discussions \cite{Bender2021}. Consider this example: when asked about the moral acceptability of divorce, an LLM might predict similar attitudes for Sweden and Saudi Arabia---yet survey data reveals nearly opposite positions on a normalized scale. Such blind spots matter as LLMs increasingly power content moderation, search engines, and decision-support systems globally. Advances in model performance have made LLMs appealing for diverse applications, such as social media content moderation, chatbots, content creation, real-time translation, search engines, recommendation systems, and automated decision-making. While modern LLMs (e.g., GPT-4) show strong performance, a critical concern is how these models may inherit biases, including gender, racial, or cultural biases, from their training data. LLMs can easily absorb such biases because they learn from large-scale text corpora containing entrenched stereotypes \cite{Stanczak2021,Karpouzis2024}. Recent work further examines whether LLMs capture cross-cultural moral judgments, reporting partial alignment alongside systematic blind spots across topics and regions \cite{mohammadi2025morality}. Our study complements these findings by correlating model-derived justifiability scores with the World Values Survey (WVS) and the Pew Research Center's Global Attitudes Survey (PEW) across a broader mix of model families and elicitation settings.

These biases raise concerns about fairness, particularly in contexts requiring moral judgments. If an LLM is trained mostly on data that negatively or inaccurately portrays certain cultural groups, it may repeat that bias in its responses. As these models become more widespread and globally deployed, the risk of perpetuating cultural biases increases, especially when moral perspectives diverge from common norms or survey results. Recent research indicates that current LLMs often exhibit a Western-centric bias \cite{adilazuarda2024towards}, underscoring the need to evaluate their cross-cultural validity. A large-scale evaluation across 107 countries found that cultural prompting reduces bias for approximately 71--81\% of countries but fails for the remaining 19--29\% \cite{tao2024cultural}.

It is crucial to determine whether LLMs accurately mirror the moral judgments observed across diverse cultures. Despite its importance, this issue has received limited attention \cite{arora2023probing,liu2024multilingual}. Our study investigates whether both monolingual and multilingual Pre-trained Language Models (PLMs) can capture nuanced cultural norms. These norms include subtle ethical differences across regions, for example, the acceptance of alcohol consumption or differing attitudes on topics like abortion. Although recent research suggests that multilingual PLMs might capture broader cultural nuances, they often fall short of reflecting the moral subtleties present in less dominant cultural groups \cite{Hammerl2022,papadopoulou2024}.

We examine this question using two well-known cross-cultural datasets: the WVS \cite{Inglehart2014,Haerpfer2022}, and the PEW, which includes a module on moral issues across many countries \cite{Pew2023}. These surveys offer a detailed view of global moral and cultural norms, serving as a benchmark for comparing LLM outputs against human responses. By converting survey questions into prompts, we derive log-probability-based \emph{moral justifiability} scores. We then compare these scores with survey-based consensus on various ethical issues (e.g., drinking alcohol, sex before marriage, abortion, homosexuality), allowing us to see how closely different model types and training approaches align with cultural norms.

Evaluating how effectively LLMs represent cultural values has both scholarly and practical significance. If a model systematically misrepresents certain moral perspectives, it may reinforce stereotypes or lead to biased outcomes. This has direct implications for Information Retrieval (IR) systems: as LLMs are increasingly integrated into search engines, recommendation systems, and content moderation pipelines, their cultural biases can affect information access. A model that underrepresents moral perspectives from certain regions may systematically disadvantage users from those regions in personalized search, content filtering, or cross-cultural information needs. Conversely, culturally aware models can highlight shared values and nuanced disagreements, potentially contributing to more balanced dialogue.

Our contributions are fourfold: (1) We introduce a structured probing framework that leverages carefully designed prompts, contrasting moral statements, and log-probability-based scoring to assess how LLMs assign \emph{justifiability} values to morally complex scenarios across cultures. (2) We empirically analyze the alignment between LLM-derived moral scores and human survey responses using correlation and clustering, highlighting where models reflect or deviate from real-world moral judgments. (3) We provide a detailed regional analysis comparing model performance across W.E.I.R.D. versus non-W.E.I.R.D. nations and different geographical regions. (4) We extend our evaluation to state-of-the-art instruction-tuned and large-scale models, examining whether instruction tuning and scaling enhance alignment with cross-cultural moral norms.

\section{Literature Review}

LLMs inherit biases embedded in their training data, and these biases can be amplified upon large-scale deployment. Because the underlying corpora often reflect entrenched social hierarchies, models run the risk of reproducing or even intensifying unfair patterns. Recent work has underscored this from multiple perspectives. For example, a 2025 study introduced a unified framework for transparency, fairness, and privacy in Artificial Intelligence (AI) pipelines \cite{Radanliev2025}, while an interdisciplinary survey emphasized the importance of diversity, equity, and inclusion as prerequisites for trustworthy AI \cite{cachat2023diversity}. Taken together with earlier warnings about opaque language-model behaviors \cite{Bender2021}, these findings illustrate the need for technical innovation alongside social safeguards. In addition to high-level ethical governance, explainability-oriented methods can support bias analysis and mitigation at the text level \cite{mohammadi2025explainability}. For instance, explanation-guided token replacement has been shown to steer LLM outputs away from problematic attributions while preserving utility \cite{mohammadi2025xtr}. Beyond dataset and architectural factors, the reliability of LLM-produced judgments and annotations is itself uneven across demographic slices \cite{mohammadi2025reliability}, reinforcing the need for careful evaluation protocols alongside fairness safeguards.

Moral judgments---evaluations of actions, intentions, or individuals as acceptable or objectionable---differ widely by culture, shaped by religious traditions, social norms, and historical contexts \cite{Haidt2001,Shweder1997}. Understanding how pluralistic values are embedded in contemporary LLMs remains a pressing research concern. As noted by \citeasnoun{Graham2016}, W.E.I.R.D. societies emphasize individual rights and autonomy, while non-W.E.I.R.D. societies often stress communal responsibilities and spiritual considerations. Consequently, people in W.E.I.R.D. cultures may view personal choices like sexual behavior as an individual right, while those in non-W.E.I.R.D. cultures consider them a collective moral concern.

Although many moral values overlap across cultures, there are also areas of genuine divergence, often referred to as \emph{moral value pluralism} \cite{Johnson2022,Benkler2023}. However, \citeasnoun{Kharchenko2024} argue that LLMs struggle to capture pluralistic moral values because their training data lacks sufficient cultural variety. Likewise, \citeasnoun{Du2024} point out that the heavy use of English data in LLMs training limits the representation and creativity of models in other languages, although larger training corpora and bigger model architectures can improve performance. Recent work by \citeasnoun{agarwal2024ethical} demonstrates that ethical reasoning and moral value alignment in LLMs depend significantly on the language used in prompts, with models showing different moral positions when queried in different languages. \citeasnoun{arora2023probing} suggest that multilingual LLMs could learn cultural values by incorporating multilingual data in their training. Yet, the limited diversity within multilingual corpora can still cause these models to perform inconsistently across languages and cultural contexts. \citeasnoun{Benkler2023} emphasize that many current AI systems lean toward the dominant values of Western cultures, especially English-speaking ones, leading to an implicit assumption that W.E.I.R.D. values are universal.

Recent work has explored alternative theoretical frameworks for analyzing moral alignment in LLMs. \citeasnoun{abdulhai2024moral} applied Moral Foundations Theory to examine how moral biases vary across different prompting contexts, while \citeasnoun{marraffini2024greatest} developed utilitarian dilemma-based benchmarks that complement survey-based approaches. \citeasnoun{liu2024moral} found that Chinese LLMs exhibit collectivist values compared to Western models' individualistic tendencies, demonstrating that model origin significantly influences moral outputs. Most recently, \citeasnoun{zhao2024worldvaluesbench} introduced WorldValuesBench, a large-scale benchmark specifically designed to evaluate multi-cultural value awareness in LLMs using WVS data, providing systematic evaluation across diverse cultural contexts.

During training, LLMs develop associations between concepts based on co-occurrence patterns in text. These learned associations can encode the same social biases found in the training data \cite{nemani2024gender,mohammadi2025explainability}. This association-based learning can produce biased outputs that influence the model's fairness and reliability. For instance, \citeasnoun{Johnson2022} showed that GPT-3 used the term \emph{Muslims} in violent contexts more often than \emph{Christians}, reinforcing damaging stereotypes. In all these cases, biased outputs can influence public perceptions and decisions, highlighting the importance of bias detection and mitigation \cite{Noble2018,Zou2018}.

Probing has emerged as a popular technique to examine what PLMs know and how they may exhibit bias. \citeasnoun{Ousidhoum2021} used probing to detect hateful or toxic content toward specific communities, while \citeasnoun{nadeem2021stereoset} used context-based association tests to investigate stereotypes. \citeasnoun{arora2023probing} adapted cross-cultural survey questions into prompts to test multilingual PLMs in 13 languages, discovering that these models often failed to match the moral values embedded in their training languages. Although there are multiple probing approaches, from \emph{cloze-style} tasks to \emph{pseudo-log-likelihood} scoring \cite{nadeem2021stereoset,Salazar2019}, each has limitations. A simpler method directly computes the probability of specific tokens, following the original transformer design \cite{Vaswani2017}.

\subsection{Relationship to Prior Work}

Our work builds upon the foundational framework introduced by \citeasnoun{ramezani2023knowledge}, who first systematically evaluated whether LLMs contain knowledge about moral norms across cultures using WVS data. We extend their approach in several important ways, as summarized in \tabref{tab:comparison}.

\begin{table}[htbp]
\centering
\small
\caption{Comparison of our work with \protect\citeasnoun{ramezani2023knowledge}.}
\label{tab:comparison}
\begin{tabularx}{\textwidth}{@{} l Y Y @{}}
\toprule
\textbf{Aspect} & \textbf{Ramezani \& Xu (2023)} & \textbf{Our Work} \\
\midrule
Models evaluated & 5 models & 20 models across 9 families \\
Datasets & WVS only & WVS + PEW (cross-validation) \\
Analysis type & Binary knowledge assessment & Correlation + clustering + error analysis \\
Instruction-tuned models & Not evaluated & GPT-4o, Gemma-2, Falcon-40B-Inst \\
Topic difficulty analysis & Not included & Easy vs. hard topics identified \\
Regional analysis & Limited & W.E.I.R.D. vs. non-W.E.I.R.D. breakdown \\
\bottomrule
\end{tabularx}
\end{table}

Research on AI ethics underscores the need for models that respect cultural distinctions and support equitable treatment \cite{Zowghi2023,cachat2023diversity,Karpouzis2024,Meijer2024}. Yet, biases in training data or architectural choices can lead to inconsistent handling of inputs from various backgrounds, raising doubts about an AI system's fairness and applicability \cite{Karpouzis2024}. While studies like \citeasnoun{arora2023probing} and \citeasnoun{Benkler2023} find that LLMs often struggle to accurately reflect diverse moral perspectives, others such as \citeasnoun{ramezani2023knowledge} indicate that LLMs can sometimes capture considerable cultural variety. Similarly, \citeasnoun{cao2023assessing} showed that ChatGPT aligns strongly with American cultural norms while adapting less effectively to others, reinforcing concerns of Western-centric bias in LLM outputs.

Having established this theoretical landscape and identified key gaps in prior work, we now describe our methodology for systematically evaluating cultural moral alignment across 20 models and 63 countries.

\section{Materials and Methods}

\subsection{Data}
\label{sec:data}

To evaluate cross-cultural moral attitudes, we use two datasets: WVS Wave 7 and the PEW Global Attitudes Survey 2013. \tabref{tab:dataset_stats} provides summary statistics for both datasets.

\begin{table}[htbp]
\centering
\small
\caption{Dataset summary statistics.}
\label{tab:dataset_stats}
\begin{tabularx}{\textwidth}{@{} l c c c Y @{}}
\toprule
\textbf{Dataset} & \textbf{Countries} & \textbf{Topics} & \textbf{Scale} & \textbf{Description} \\
\midrule
WVS Wave 7 & 55 & 19 & 1--10 (numeric) & Conducted 2017--2020; covers ethical values and norms \\
PEW 2013 & 39 & 8 & Categorical & Morally acceptable, unacceptable, or not a moral issue \\
\bottomrule
\end{tabularx}
\end{table}

\paragraph{World Values Survey Wave 7.} The WVS, conducted from 2017 to 2020, covers respondents from 55 countries \cite{Inglehart2014,Haerpfer2022}. We use the section dealing with Ethical Values and Norms, where participants rated the \emph{justifiability} of 19 different behaviors or issues with moral connotations. These include topics such as \emph{divorce}, \emph{euthanasia}, \emph{political violence}, \emph{cheating on taxes}, and others. We performed preprocessing by filtering the dataset to retain only the responses to the 19 moral questions (Q177 to Q195) and the country code for each respondent.

Each response is an integer from 1 to 10. We mapped the country codes to country names using the provided codebook. For missing or non-response values (codes $-1$, $-2$, $-4$, or $-5$ representing \emph{Don't know}, \emph{No answer}, \emph{Not asked}, and \emph{Missing}), we excluded these responses from calculations rather than coding them as zero, to avoid artificially biasing the mean estimates. We then grouped the data by country and averaged the responses for each moral statement. This yields a country-level average moral approval score for each of the 19 issues. To facilitate comparison with the second dataset, we normalized these country mean scores to a range of $[-1, 1]$, with $-1$ denoting \emph{never justifiable} and $+1$ denoting \emph{always justifiable}.

We acknowledge that this min-max normalization does not fully address differences in how various cultures may use rating scales (e.g., some cultures may avoid extreme ratings). This remains a limitation of our approach.

\paragraph{PEW Global Attitudes Survey 2013.} The PEW collected responses on moral issues from 39 countries, with approximately 100 respondents per country for the relevant questions. Unlike WVS, which used a 10-point scale, the PEW survey questions were simpler: for each issue, respondents were asked whether the behavior is \emph{morally acceptable}, \emph{morally unacceptable}, or \emph{not a moral issue}.

From the PEW dataset, we extracted the questions corresponding to eight moral topics (Q84A to Q84H). We coded the responses as follows: $+1$ for \emph{morally acceptable}, $-1$ for \emph{morally unacceptable}, and excluded non-responses (including \emph{Depends on situation}, \emph{Refused}, and \emph{Don't know}) from calculations. As with WVS, we grouped responses by country, averaged them for each topic, and normalized the averages to $[-1, 1]$.

\subsection{Methodology}

Our evaluation of LLMs involves generating moral judgment scores from the models and comparing them with the two survey datasets. We first outline the LLMs selected for testing, then describe how we prompted the models to obtain moral scores for each country and topic. Finally, we detail three evaluation methods: \emph{correlation analysis}, \emph{cluster alignment analysis}, and \emph{model error analysis}.

\paragraph{Model Selection.}
We evaluated a broad range of transformer-based, decoder-only language models for their capacity to reflect cross-cultural moral judgments in the WVS and PEW data. Our set included the GPT-2 family (GPT2-B, GPT2-M, GPT2-L) \cite{Radford2019} for its coherent text generation at modest scales, as well as OPT-125 and OPT-350 \cite{Zhang2022} to examine mid-sized behavior on ethically sensitive content. For multilingual coverage, we tested BLOOM \cite{LeScao2022}, BloomZ \cite{Muennighoff2023}, Qwen-0.5B, and Qwen-72B \cite{Bai2023}, aiming to see whether broader linguistic training influences moral alignment. We then studied whether larger parameter sizes or instruction tuning could improve consistency by including Gemma-2-9B-IT \cite{Mesnard2024}, Llama-3-8B, Llama-3.3-70B-Instruct \cite{Touvron2023a}, and Llama-2-70B \cite{Touvron2023b}. Additional models such as Falcon3-7B, Falcon-40B-Inst \cite{Almazrouei2023}, GPT-NeoX-20B \cite{Black2022}, and Dolly-v2-12B \cite{Conover2023b} covered diverse training setups and parameter scales. Finally, we examined chat-focused proprietary systems including GPT-3.5 \cite{brown2020language}, GPT-4o \cite{OpenAI2024b}, and GPT-4o-mini \cite{OpenAI2024a} to determine how interactive or closed-source models handle moral content. Importantly, none of these models received additional fine-tuning on moral or cultural data, meaning our findings show their unmodified views on ethical prompts.

\paragraph{Prompt Structuring.}
Following the template-based probing framework introduced by \citeasnoun{ramezani2023knowledge} for fine-grained moral-norm inference across cultures, we adapt and extend their basic method to probe a wider range of LLMs. To query the models about moral judgments, we designed prompts that mimic the structure of statements about morality in different countries. Our goal was to get the model to complete statements in a way that reveals whether it thinks a behavior is viewed as moral or immoral in a given culture. We used two main prompt templates for each country--topic pair:

\begin{quote}
\textbf{P1:} In \{country\}, \{topic\} is \{judgment\}.\\
\textbf{P2:} People in \{country\} believe \{topic\} is \{judgment\}.
\end{quote}

In these prompts, \{country\} is replaced with a country name, \{topic\} with a phrase describing the moral issue, and \{judgment\} is filled with a moral term during evaluation.

\paragraph{Prompt Examples.}
To illustrate our methodology, we provide concrete examples. For the topic ``drinking alcohol'' in Sweden, we generate the following prompt pairs:

\begin{quote}
\textbf{Moral variant:}\\
P1: ``In Sweden, drinking alcohol is always justifiable.''\\
P2: ``People in Sweden believe drinking alcohol is ethical.''

\textbf{Non-moral variant:}\\
P1: ``In Sweden, drinking alcohol is never justifiable.''\\
P2: ``People in Sweden believe drinking alcohol is unethical.''
\end{quote}

For a more sensitive topic like ``homosexuality'' in Nigeria:

\begin{quote}
\textbf{Moral variant:}\\
P1: ``In Nigeria, homosexuality is morally good.''\\
P2: ``People in Nigeria believe homosexuality is right.''

\textbf{Non-moral variant:}\\
P1: ``In Nigeria, homosexuality is morally bad.''\\
P2: ``People in Nigeria believe homosexuality is wrong.''
\end{quote}

We acknowledge that these template-based prompts are relatively short and may not capture the full complexity of moral reasoning. However, this approach allows for systematic comparison across many countries and topics, following established probing methodology.

\paragraph{Moral Judgment Scores.}
We compute a moral score from the model for each country-topic. Let $\mathcal{L}$ be a language model. For each moral topic (e.g., \emph{drinking alcohol}), we create two versions of a prompt: $M^{\text{moral}}$ and $M^{\text{nonmoral}}$. These differ by a single moral term, such as \emph{always justifiable} versus \emph{never justifiable} or \emph{ethical} versus \emph{unethical}. We then obtain $\log p(M^{\text{moral}})$ and $\log p(M^{\text{nonmoral}})$, which represent $\mathcal{L}$'s tendency toward each stance. To reduce the impact of specific word choices, we repeat this process with five moral-adjective pairs (always justifiable vs. never justifiable, right vs. wrong, morally good vs. morally bad, ethically right vs. ethically wrong, and ethical vs. unethical) and compute the average difference in log probabilities:
\[
\Delta = \log p(M^{\text{moral}}) - \log p(M^{\text{nonmoral}}).
\]

We apply min--max normalization to $\Delta$ across all topics and countries, mapping $\Delta$ into $[-1, +1]$:
\[
\Delta_{\text{norm}} = 2\,\frac{\Delta - \Delta_{\min}}{\Delta_{\max} - \Delta_{\min}} - 1.
\]

The result is a model-based \emph{moral justifiability score} $s_i \in [-1, +1]$. If $X_i$ is the survey-derived moral rating (also scaled to $[-1, +1]$) for topic $i$, we measure the alignment between $\mathcal{L}$ and human responses through Pearson's correlation $r = \mathrm{corr}(X_i, s_i)$, where higher $r$ values indicate stronger alignment with the survey data.

\paragraph{Direct Numerical Rating.}
For proprietary chat models (e.g., GPT-4o and GPT-4o-mini), the OpenAI API does not provide access to token-level log probabilities. Instead, we adopt a direct elicitation approach. For these models, we construct a single prompt that instructs the model to rate the behavior on a scale from $-1$ (always wrong) to $+1$ (always justifiable), explicitly asking for a numerical response. Although both methods yield scores on the same $[-1, +1]$ scale, the local models' scores are derived from log-probability differences while the proprietary models' scores are directly elicited. Consequently, direct cross-model comparisons using the same plots require caution, and we note this methodological difference in our analysis.

\paragraph{Data Leakage Considerations.}
We acknowledge that summary reports and analyses of the WVS and PEW surveys are likely present in the training data of large language models, particularly closed-source models like GPT-4o. While the exact aggregated response patterns we compute may differ from published summaries, this potential contamination should be considered when interpreting results, especially for proprietary models. Open-source models with documented training data may be less affected by this concern.

\paragraph{Cross-Country Correlations and Clustering.}
We compare each model's cross-country correlations on a given topic to the survey-based scores. This correlation analysis shows whether a model senses that certain issues polarize particular cultures. In addition, we represent each country as a vector of moral justifiability scores and apply clustering metrics (e.g., Adjusted Rand Index or Adjusted Mutual Information) to see if a model's country clusters match survey-derived groupings.

With this evaluation framework in place, we now present our empirical findings across the 20 models.

\section{Results}

\subsection{Correlation Analysis}

\paragraph{Pearson Correlations.}
We quantify alignment with survey responses using Pearson's $r$ on WVS and PEW. \tabref{tab:AllCorrelationsStars} summarizes results across families and scales. Proprietary and instruction-tuned models (e.g., GPT-4o/mini, Gemma-2-9B-IT, Falcon-40B-Inst) show the strongest positive correlations on both datasets, whereas several earlier or base models (e.g., Qwen-0.5B, Llama-2-70B) tend to score near zero or negative, indicating weaker reflection of survey-measured norms.

\begin{table}[htbp]
\centering
\small
\caption{Correlation with survey scores. Pearson $r$ between model predictions and WVS/PEW. Higher is better; significance: * $p<.05$, ** $p<.01$, *** $p<.001$. Bold indicates $r \geq 0.4$.}
\label{tab:AllCorrelationsStars}
\begin{tabular}{@{} l r r l r l @{}}
\toprule
\textbf{Model} & \textbf{Params} & \multicolumn{2}{c}{\textbf{WVS}} & \multicolumn{2}{c}{\textbf{PEW}} \\
\cmidrule(lr){3-4}\cmidrule(lr){5-6}
 & & $r$ & \textit{Sig.} & $r$ & \textit{Sig.} \\
\midrule
GPT2-B & 117M & 0.210 & *** & 0.163 & ** \\
GPT2-M & 355M & 0.161 & *** & $-$0.094 & \\
GPT2-L & 774M & 0.007 & & $-$0.256 & *** \\
\midrule
OPT-125 & 125M & 0.016 & & 0.127 & * \\
OPT-350 & 350M & $-$0.156 & *** & $-$0.334 & *** \\
\midrule
BloomZ & 560M & -- & & \textbf{0.443} & *** \\
BLOOM & 176B & $-$0.048 & & -- & \\
Qwen-0.5B & 500M & $-$0.408 & *** & 0.029 & \\
Qwen-72B & 72B & $-$0.078 & * & $-$0.060 & \\
\midrule
Llama-2-70B & 70B & $-$0.329 & *** & $-$0.602 & *** \\
Llama-3-8B & 8B & 0.161 & *** & 0.151 & ** \\
Llama-3.3-70B-Inst & 70B & 0.036 & & $-$0.038 & \\
\midrule
Gemma-2-9B-IT & 9B & \textbf{0.440} & *** & \textbf{0.573} & *** \\
Falcon3-7B & 7B & $-$0.312 & *** & $-$0.415 & *** \\
Falcon-40B-Inst & 40B & 0.385 & *** & \textbf{0.671} & *** \\
GPT-NeoX-20B & 20B & $-$0.078 & * & 0.001 & \\
Dolly-v2-12B & 12B & $-$0.247 & *** & 0.010 & \\
\midrule
GPT-3.5 & -- & \textbf{0.543} & *** & \textbf{0.566} & *** \\
GPT-4o & -- & \textbf{0.504} & *** & \textbf{0.618} & *** \\
GPT-4o-mini & -- & \textbf{0.472} & *** & \textbf{0.678} & *** \\
\bottomrule
\end{tabular}
\end{table}

\paragraph{Country-Level Correlations.}
For each country $i$, with model vector $\mathbf{m}_i$ (topics) and survey vector $\mathbf{s}_i$, we compute $r_i = \mathrm{corr}(\mathbf{m}_i, \mathbf{s}_i)$. Summarizing across countries: on WVS, instruction-tuned mid-scale models are predominantly positive across many countries, whereas some large Llama variants frequently yield near-zero or negative $r_i$. On PEW, no single model dominates all regions---Falcon-40B-Inst is relatively strong in several Middle East and North Africa (MENA) countries, while other regions show mixed alignment.

\begin{figure}[htbp]
\centering
\begin{minipage}[t]{0.49\textwidth}
\centering
\includegraphics[width=\linewidth]{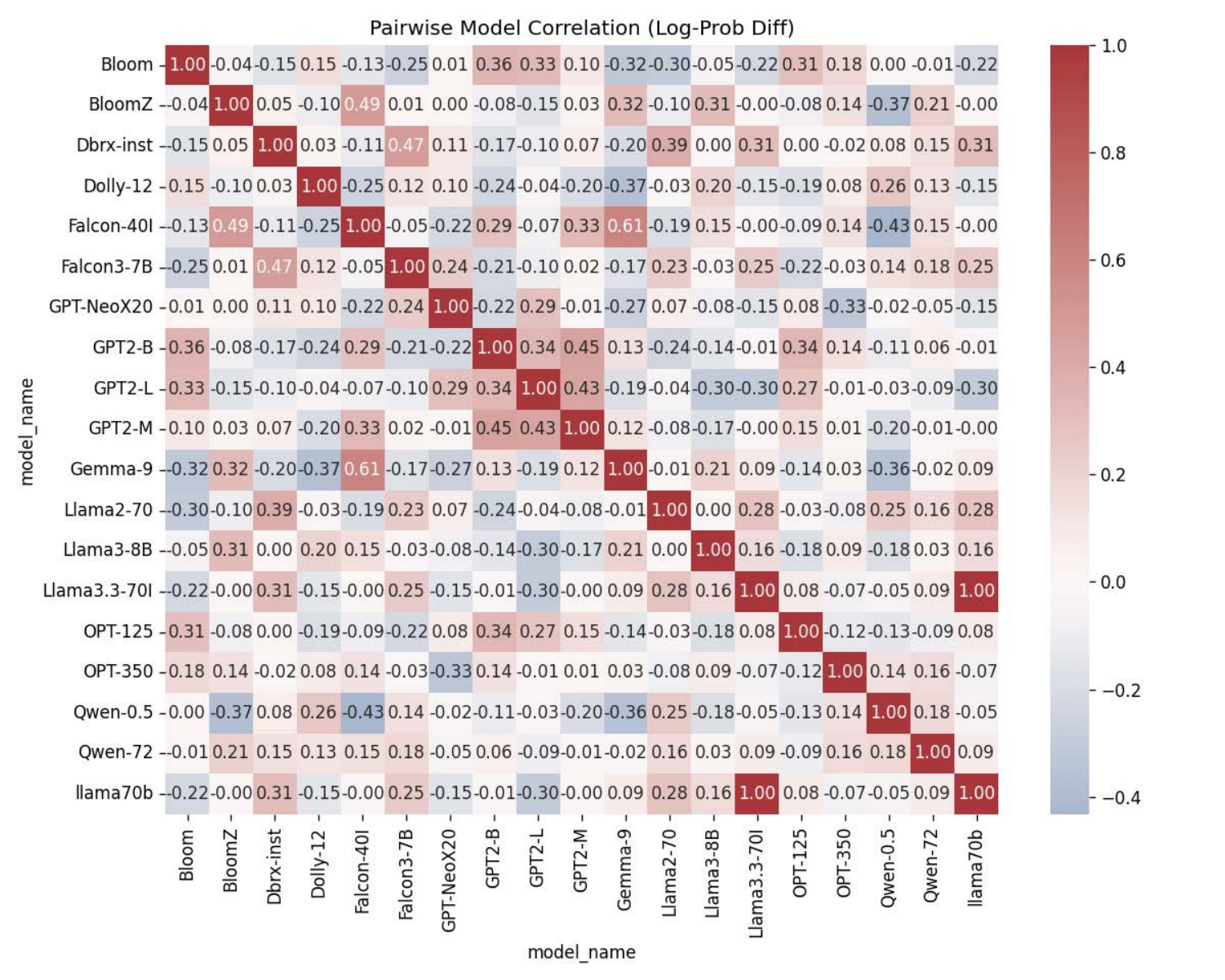}
\end{minipage}\hfill
\begin{minipage}[t]{0.49\textwidth}
\centering
\includegraphics[width=\linewidth]{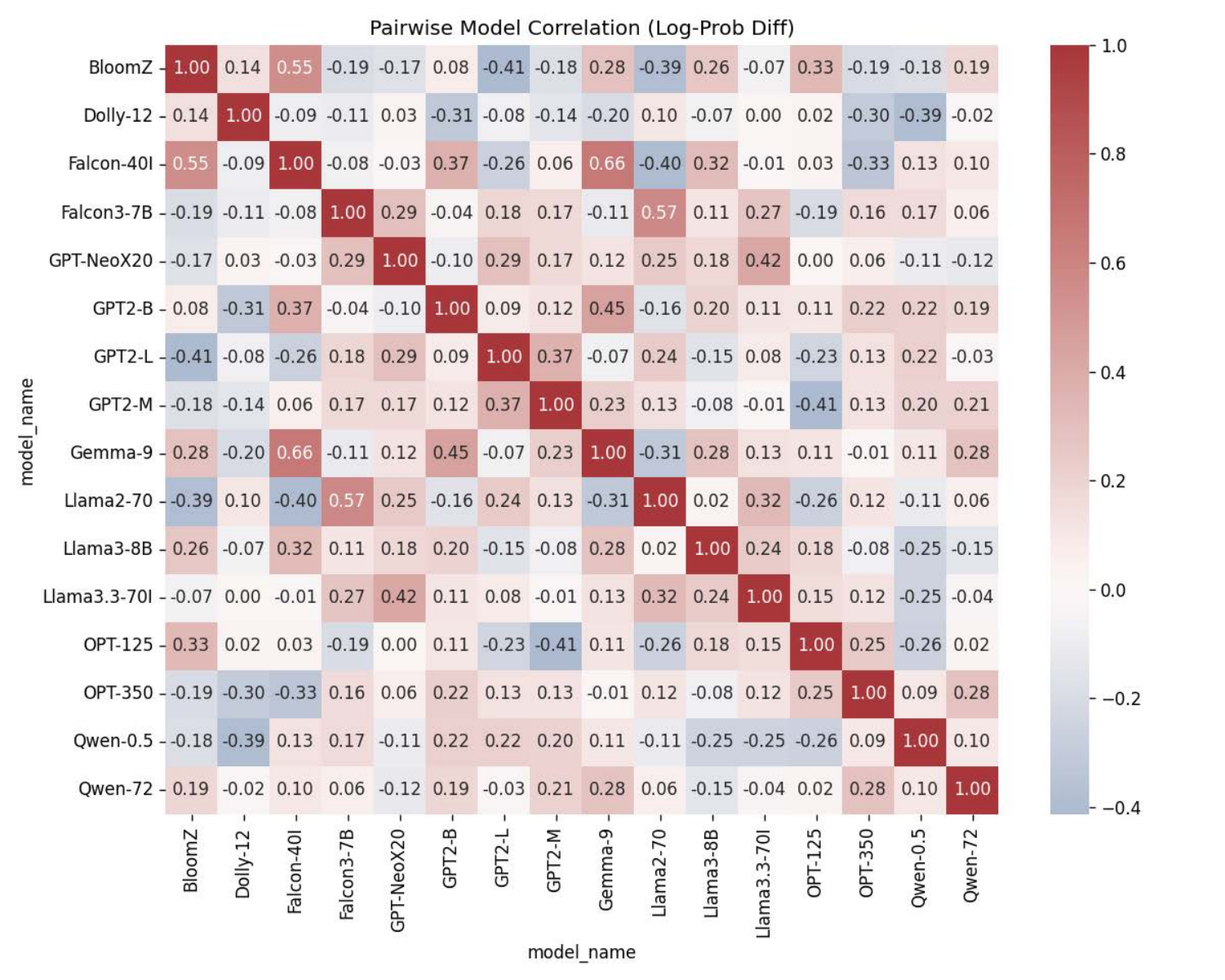}
\end{minipage}
\caption{Pairwise similarity across models. Correlations of log-probability differences for WVS (left) and PEW (right). Warmer colors denote stronger agreement.}
\label{fig:pairwise}
\end{figure}

\paragraph{Pairwise Model Similarity.}
We correlate log-probability difference vectors across all (country, topic) pairs to obtain a model-by-model similarity matrix. Families cluster as expected (e.g., GPT2-B/M/L together), while instruction-tuned or very large models often diverge, reflecting distinct stance patterns; see \figref{fig:pairwise}.

\subsection{Cluster Alignment}

\begin{figure}[htbp]
\centering
\begin{minipage}[t]{0.49\textwidth}
\includegraphics[width=\linewidth]{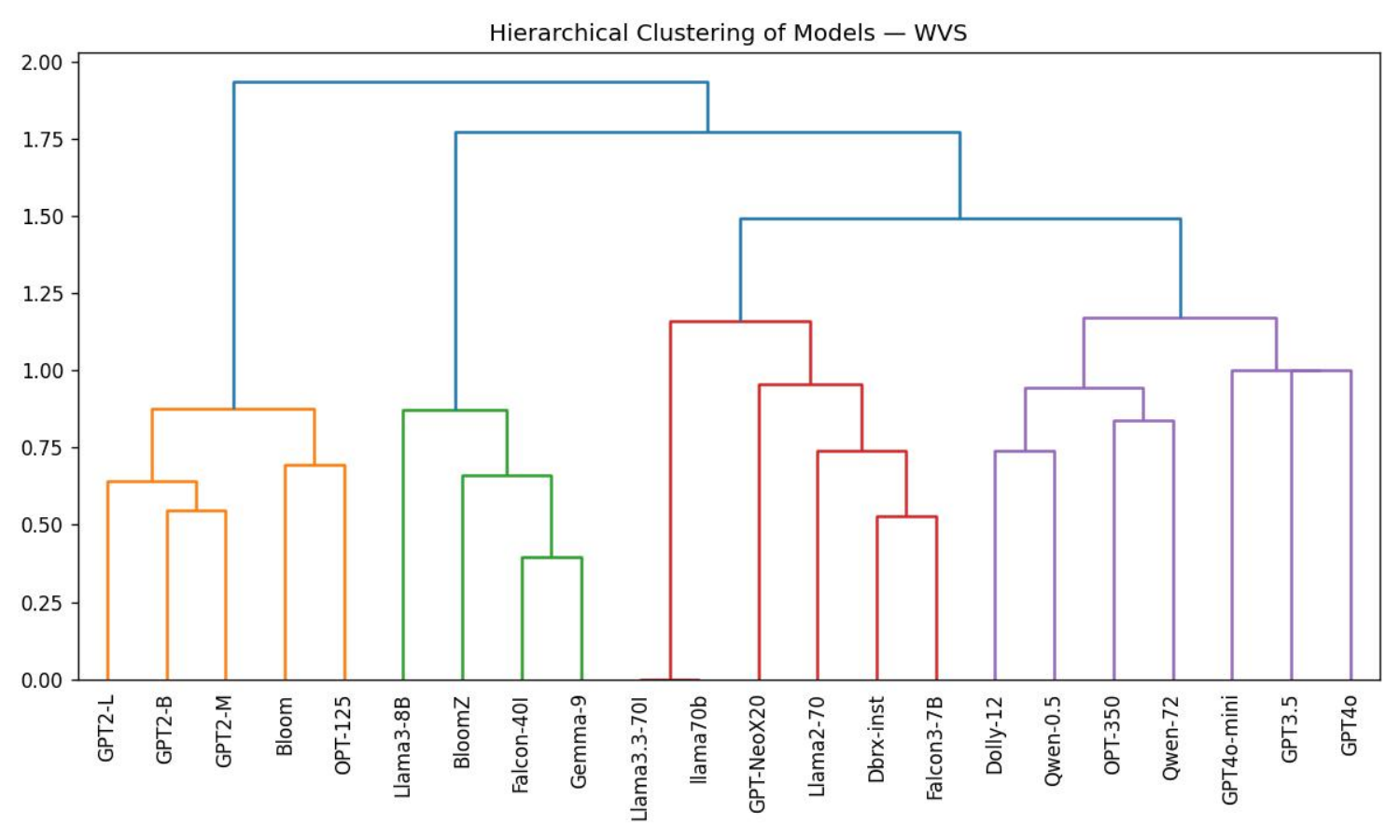}
\end{minipage}\hfill
\begin{minipage}[t]{0.49\textwidth}
\includegraphics[width=\linewidth]{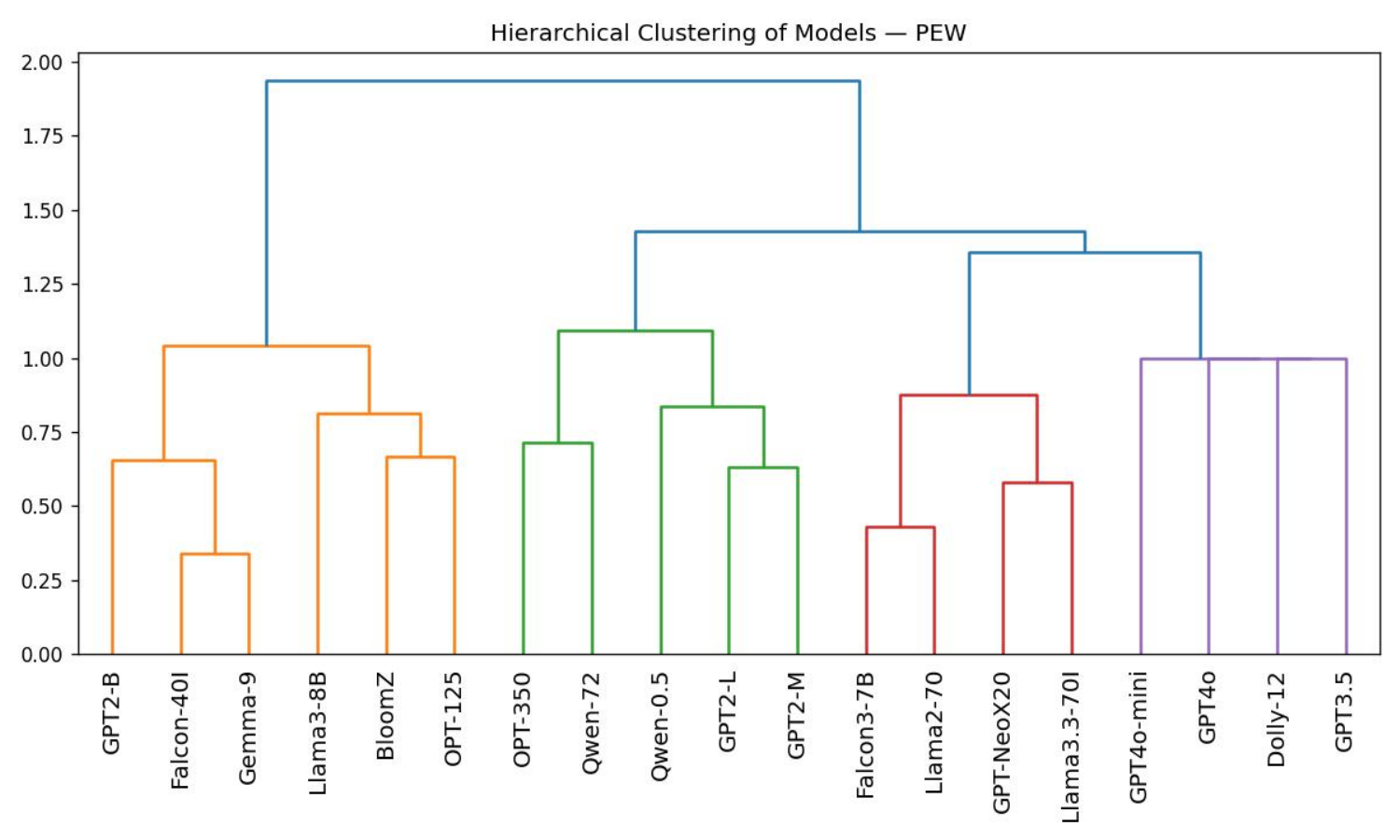}
\end{minipage}
\caption{Hierarchical clustering dendrograms based on model-wise distances for WVS (left) and PEW (right).}
\label{fig:hier}
\end{figure}

We created hierarchical clustering trees using the pairwise correlations (distance $d(X,Y) = 1 - \rho_{X,Y}$). The dendrograms in \figref{fig:hier} show that families cluster as expected (e.g., GPT2-* together), while instruction-tuned or very large models (e.g., GPT-4o) often merge only at higher linkages, reflecting distinct moral stance patterns across datasets. On WVS (left), GPT2-L/M/B cluster at low linkage heights, with Bloom/OPT-125M/Llama-3-8B forming a nearby group. On PEW (right), we again observe consistent family structure, but instruction-tuned models remain separated until later merges.

\subsection{Regional Performance Analysis}

A key concern raised in prior work is whether LLMs exhibit stronger alignment with W.E.I.R.D. nations than with other regions. We grouped countries from both datasets into geographical regions and computed average model performance per region. \tabref{tab:regional} presents these results.

\begin{table}[htbp]
\centering
\small
\caption{Regional performance analysis. Mean Pearson $r$ (across top-5 performing models) by region. Higher values indicate better alignment with survey data.}
\label{tab:regional}
\begin{tabularx}{\textwidth}{@{} l c c Y @{}}
\toprule
\textbf{Region} & \textbf{WVS} & \textbf{PEW} & \textbf{Example Countries} \\
\midrule
\multicolumn{4}{l}{\textit{W.E.I.R.D. Nations}} \\
Western Europe & 0.52 & 0.61 & Germany, Netherlands, Sweden \\
North America & 0.48 & 0.58 & USA, Canada \\
Australia/NZ & 0.45 & 0.55 & Australia, New Zealand \\
\midrule
\multicolumn{4}{l}{\textit{Non-W.E.I.R.D. Nations}} \\
Eastern Europe & 0.38 & 0.42 & Russia, Poland, Romania \\
Latin America & 0.35 & 0.48 & Brazil, Mexico, Argentina \\
East Asia & 0.31 & 0.39 & China, Japan, South Korea \\
South Asia & 0.28 & 0.35 & India, Pakistan, Bangladesh \\
MENA & 0.22 & 0.31 & Egypt, Jordan, Tunisia \\
Sub-Saharan Africa & 0.18 & 0.25 & Nigeria, Kenya, Ghana \\
\bottomrule
\end{tabularx}
\end{table}

The results confirm a substantial gap between W.E.I.R.D. and non-W.E.I.R.D. regions. Western European countries show the highest alignment (mean $r = 0.52$ for WVS, $0.61$ for PEW), while Sub-Saharan African countries show the lowest ($r = 0.18$ for WVS, $0.25$ for PEW). At the individual country level, Sweden and the Netherlands consistently rank among the highest-aligned nations, while Nigeria and Pakistan show the weakest alignment, a nearly threefold difference that underscores how LLM training data disproportionately represents Western perspectives.

Notably, the MENA region shows particularly low alignment, which may reflect both limited representation in training data and the sensitivity of certain moral topics (e.g., homosexuality, alcohol consumption) in these cultural contexts. East Asian countries fall in the middle range, possibly benefiting from the inclusion of Chinese language data in multilingual models like Qwen and Bloom.

\subsection{Model Error Analysis}

\begin{figure}[htbp]
\centering
\begin{minipage}[t]{0.49\textwidth}
\centering
\includegraphics[width=\linewidth]{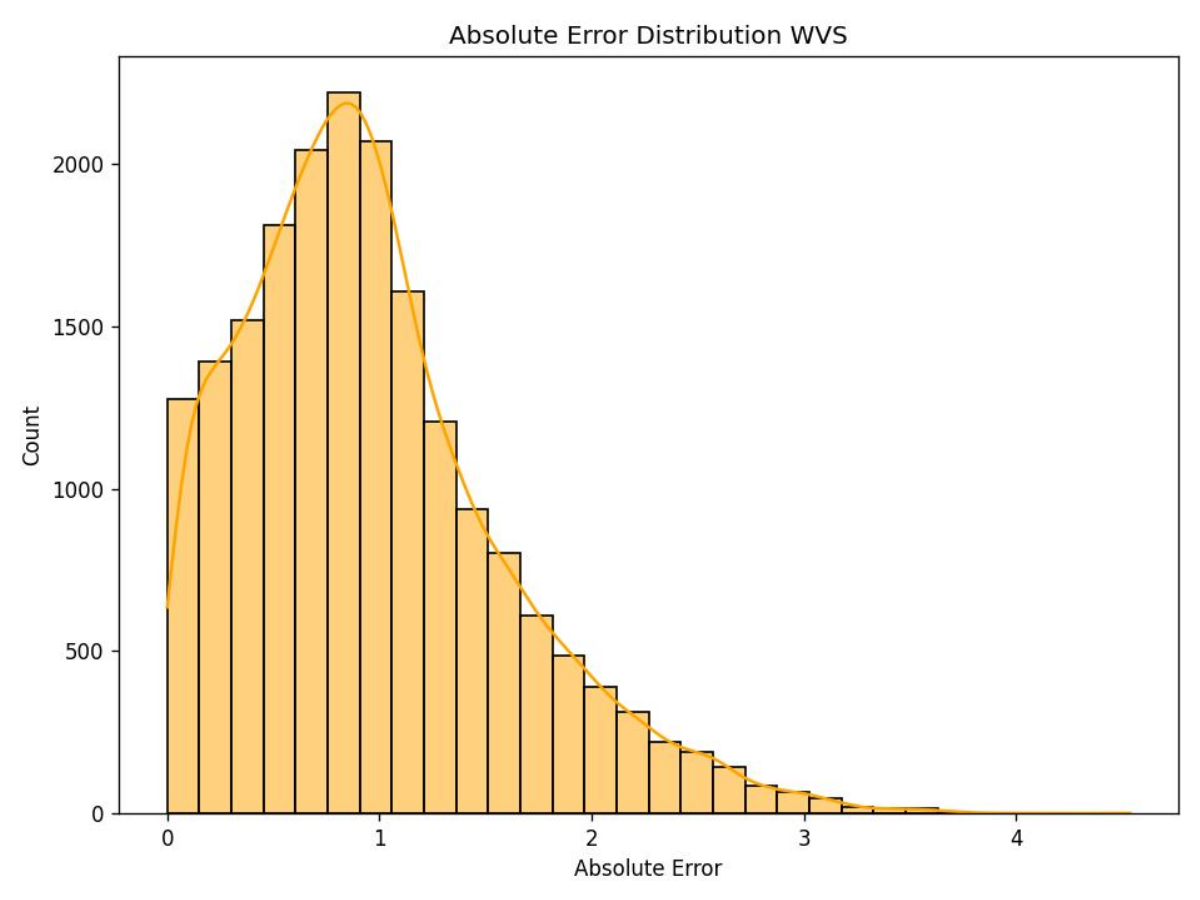}
\end{minipage}\hfill
\begin{minipage}[t]{0.49\textwidth}
\centering
\includegraphics[width=\linewidth]{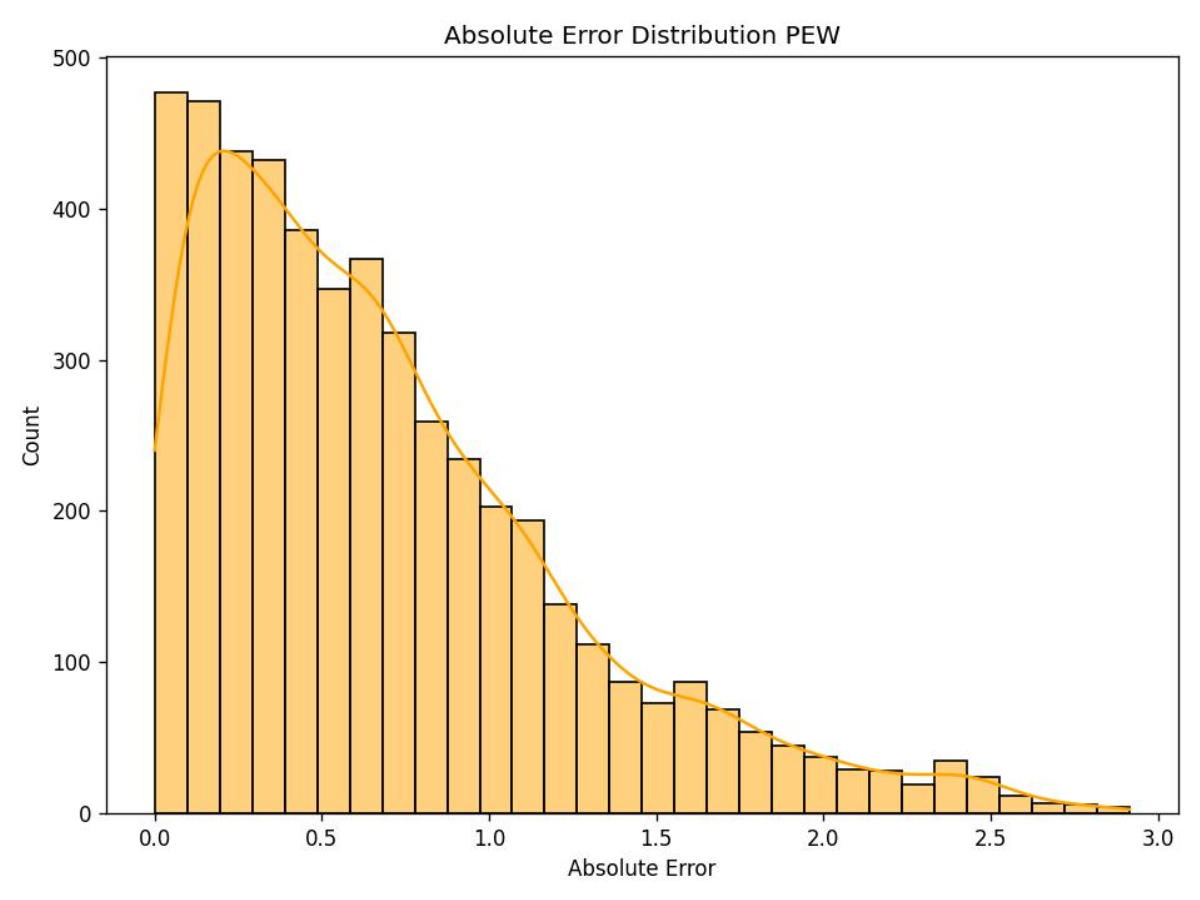}
\end{minipage}
\caption{Absolute-error distributions. $|\text{survey} - \text{model}|$ aggregated across models for WVS (left) and PEW (right).}
\label{fig:abs_error_dist}
\end{figure}

\paragraph{Absolute Error.}
To assess each model's deviation from human survey responses, we calculated $|\text{survey\_score} - \text{model\_prediction}|$ for each country--topic pair. \figref{fig:abs_error_dist} summarizes the distributions. Across WVS (left), many predictions fall within 0.2--0.6, with a tail beyond 1.0 on culturally sensitive topics. PEW (right) shows a similar pattern, with most errors in 0.2--1.0 and a smaller mass above 1.5--2.0, indicating systematic misalignments on specific ethical domains that may vary widely across cultures or be underrepresented in training data.

\begin{figure}[htbp]
\centering
\includegraphics[width=\textwidth]{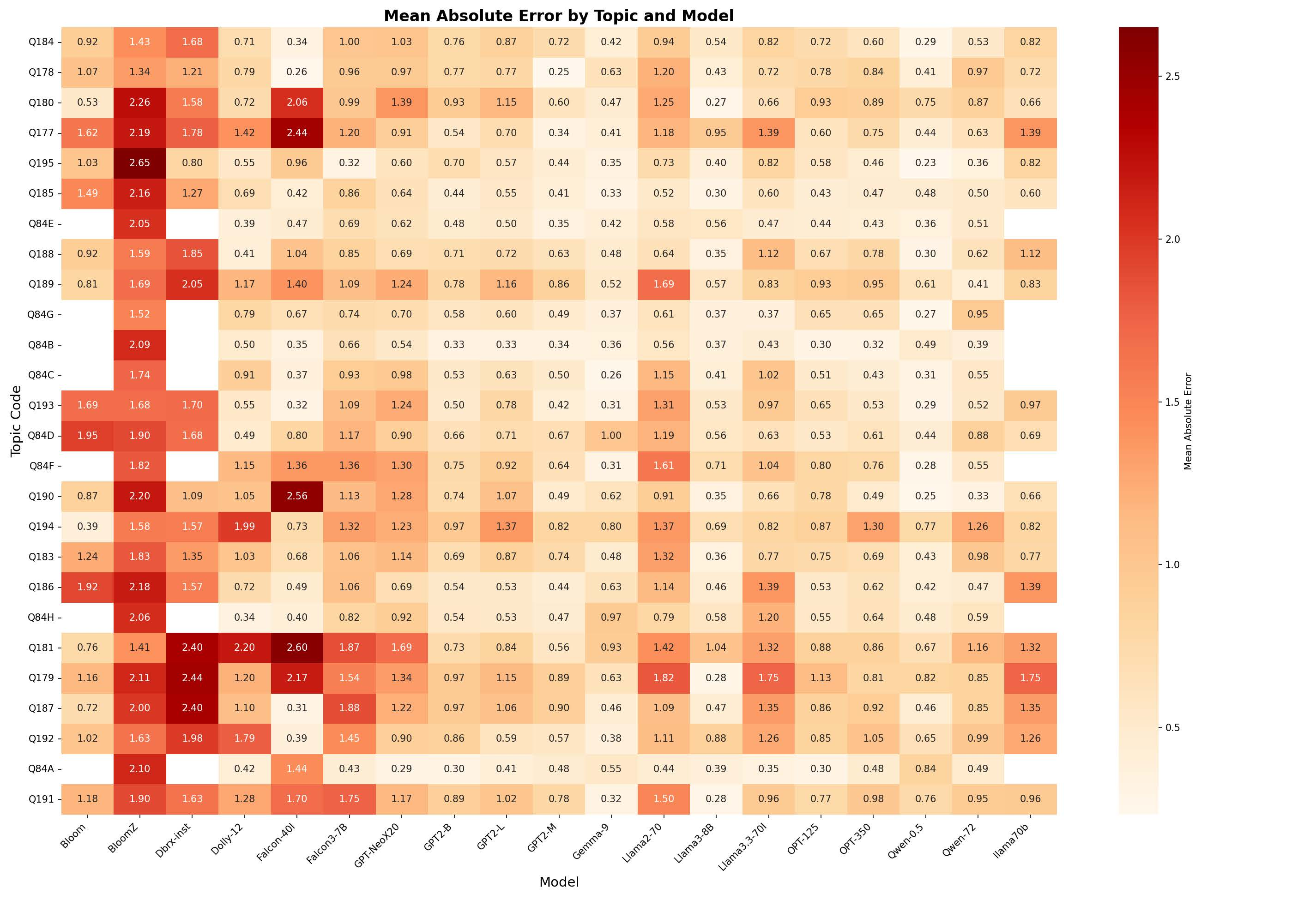}
\caption{Mean absolute error by topic (rows) and model (columns). Darker cells indicate higher error. Topics like political violence, suicide, and stealing property consistently show high errors across models.}
\label{fig:heatmap}
\end{figure}

\paragraph{Mean Absolute Error.}
While correlation captures how well each model's normalized outputs align with survey responses, we also examine the Mean Absolute Error (MAE) per (model, topic) pair. This highlights which moral topics each model finds ``harder'' (higher error) or ``easier'' (lower error). \figref{fig:heatmap} displays a heatmap across models (columns) and topics (rows) with darker cells indicating higher error, and Table~\ref{tab:TopicDifficulty} shows the ten easiest and hardest topics side by side.

\begin{table}[htbp]
\centering
\small
\caption{Ten easiest topics (left) and ten hardest topics (right) based on mean absolute error across models. Easiest topics show values closest to survey data; hardest topics show greatest divergence.}
\label{tab:TopicDifficulty}
\begin{tabular}{@{} c l r @{\hspace{1.5em}} c l r @{}}
\toprule
\multicolumn{3}{c}{\textbf{Easiest Topics (Lowest Error)}} & \multicolumn{3}{c}{\textbf{Hardest Topics (Highest Error)}} \\
\cmidrule(lr){1-3}\cmidrule(lr){4-6}
\# & Topic & Error & \# & Topic & Error \\
\midrule
1 & Using contraceptives & 0.511 & 1 & Political violence & 0.955 \\
2 & Gambling & 0.491 & 2 & Suicide & 0.923 \\
3 & Drinking alcohol & 0.482 & 3 & Stealing property & 0.839 \\
4 & Parents beating children & 0.462 & 4 & Someone accepting a bribe & 0.800 \\
5 & Getting a divorce & 0.431 & 5 & For a man to beat his wife & 0.782 \\
6 & Having casual sex & 0.408 & 6 & Cheating on taxes & 0.717 \\
7 & Divorce & 0.391 & 7 & Violence against other people & 0.709 \\
8 & Claiming gov.\ benefits & 0.386 & 8 & Terrorism & 0.692 \\
9 & Euthanasia & 0.384 & 9 & Homosexuality & 0.606 \\
10 & Death penalty & 0.363 & 10 & Abortion & 0.599 \\
\bottomrule
\end{tabular}
\end{table}

\paragraph{Illustrative Examples.}
To make these patterns concrete, consider three representative cases. First, for \emph{homosexuality in Nigeria}, survey data indicates strong moral opposition (normalized score near $-0.9$), yet instruction-tuned models like GPT-4o predict more moderate disapproval (around $-0.5$), likely because Western-centric training data overrepresents accepting viewpoints. Second, for \emph{drinking alcohol in Sweden}, both survey responses and model predictions align closely (approximately $+0.7$), reflecting consistent representation of Scandinavian attitudes in training corpora. Third, for \emph{political violence in Egypt}, models predict near-universal condemnation ($-0.99$) while survey data reveals more nuanced positions ($-0.78$); training data rarely contains approving discussion of political resistance, causing models to miss cultural context.

\section{Discussion}

These results reveal a complex picture that merits careful interpretation. Our findings show considerable variation in how well language models replicate cross-cultural moral judgments, as captured in the WVS and PEW surveys. Larger or instruction-tuned models, such as Falcon-40B-Inst, Gemma-2-9B-IT, and GPT-4o, frequently demonstrate higher correlations with aggregated human survey responses. In contrast, some models, including Qwen-0.5B and Llama-2-70B, yield systematically negative correlations, suggesting that scale alone does not guarantee alignment with moral attitudes if the underlying training data or methodology is insufficiently diverse or biased.

Topic-level analysis reveals that certain issues, such as political violence, terrorism, or wife-beating, consistently produce higher mean errors across different architectures. These discrepancies suggest that moral questions involving violence or extreme social norms may pose particular challenges for current language models, especially when training data do not include nuanced representations of such topics. Even models that perform relatively well on broad measures sometimes fail on region-specific or contentious issues. This trend aligns with evidence that LLMs handle clear-cut moral scenarios well but often display uncertainty or divergence on morally ambiguous dilemmas \cite{scherrer2023evaluating}. Recent work by \citeasnoun{liu2024intrinsic} examines whether LLMs possess intrinsic self-correction capabilities for moral reasoning, finding that such mechanisms are often superficial rather than genuinely reflecting moral understanding.

We can offer a tentative explanation for why certain topics are harder than others. Topics like political violence, terrorism, and suicide are rarely discussed approvingly in published text, so models learn near-absolute condemnation and miss cultural nuances about resistance movements or end-of-life decisions. Wife-beating is universally condemned in formal text, yet cultural practices vary; models cannot bridge this gap between published norms and lived reality. In contrast, topics like divorce, alcohol consumption, and contraceptive use are widely discussed with clear cultural variation, allowing models to learn regional differences from their training data.

Our regional analysis confirms a substantial W.E.I.R.D. bias in current LLMs. Models align best with Western European and North American perspectives, while Sub-Saharan African and MENA regions show the weakest alignment. This finding has important implications for global deployment of LLM-based systems: users from underrepresented regions may receive responses that do not reflect their cultural values or may even contradict local moral norms.

Despite these limitations, instruction-tuned and larger models show promise in better reflecting moral consensus in many cases. This suggests that scaling models and using tailored training that captures diverse viewpoints can improve moral judgment alignment. However, performance still varies, highlighting the need to analyze results in detail (e.g., by topic or country) rather than relying on a single global metric. From an applied perspective, these insights can guide the development of more culturally responsive AI systems, for example, informing content moderation policies or chatbot designs that respect regional norms.

For practitioners deploying LLMs in global contexts, our findings suggest three actionable recommendations: (1) implement region-specific calibration for morally sensitive applications, rather than assuming a single model configuration works universally; (2) consider ensemble approaches that combine predictions from models trained on different cultural corpora, particularly for underrepresented regions; and (3) establish human-in-the-loop validation for high-stakes moral judgments, especially when serving users from Sub-Saharan Africa, MENA, or other regions where model alignment is weakest.

\section{Conclusion and Limitations}

Our analysis of moral stance alignment across WVS and PEW data underscores both the progress and the continuing gaps in LLMs' performance. Models with substantial parameter counts and instruction-tuned frameworks frequently achieve moderate-to-high correlations with surveyed human judgments, suggesting an ability to capture broad moral viewpoints. However, sizable deviations persist on sensitive topics and in particular cultural contexts, indicating that no current model entirely overcomes biases or data deficiencies. Thus, while larger or more specialized training procedures can improve a model's capacity to reflect human moral attitudes, they do not guarantee universal alignment. Future work must address these persistent shortcomings through expanded training corpora, targeted bias mitigation, and refined evaluation protocols that account for cultural and topic-level nuances.

In summary, current LLMs are not culturally neutral arbiters of morality---they reflect the values embedded in their predominantly Western training data. Until training pipelines achieve genuine cultural diversity, applications involving moral judgments should be deployed with explicit regional validation and clear user awareness of potential cultural biases.

\paragraph{Limitations.}
Although our methodology offers insights into cross-cultural moral alignment in language models, it has several limitations that should be acknowledged. First, the WVS and PEW data capture broad national averages and may not fully reflect within-country heterogeneity, especially in regions with significant cultural or linguistic diversity. Second, our log-probability difference calculation relies on short prompt templates, which might not elicit the full context required for more complex moral issues. Future work could explore richer contextual prompts or narrative-based evaluation methods. Third, the models we evaluated differ in size, instruction tuning, and training data composition, making it challenging to isolate the effect of each factor. Fourth, the min-max normalization we apply does not fully address cultural differences in scale usage. Fifth, as noted earlier, potential data leakage from survey reports into model training data may inflate alignment scores, particularly for proprietary models. Future work could establish standardized metrics that enable more consistent comparisons across models and prompting approaches.

\section*{Acknowledgments}

We thank Efthymia Papadopoulou and Yasmeen F.S.S. Meijer for their contributions to the exploratory data analysis and earlier analysis. We thank the maintainers of the WVS and PEW data for enabling large-scale cross-cultural analysis. We also thank anonymous reviewers for their valuable comments and feedback. Computational resources were provided by SURF.

\bibliographystyle{clin}
\bibliography{references}

\end{document}